\documentclass{winnower}
\usepackage[utf8]{inputenc} 
\usepackage[T1]{fontenc}    
\usepackage{hyperref}       
\usepackage{url}            
\usepackage{booktabs}       
\usepackage{amsfonts}       
\usepackage{nicefrac}       
\usepackage{microtype}      
\usepackage{latexsym}
\usepackage{mathtools}
\usepackage{algorithm}
\usepackage{algorithmic}
\usepackage{multirow}
\usepackage{natbib}
\usepackage{graphicx}
\usepackage{epstopdf}
\setcitestyle{numbers}
\begin{document}
\lhead[]{}
\title{Towards Crafting Text Adversarial Samples}

\author{Suranjana Samanta}
\affil{IBM India Research Lab (IRL), Bangalore, India. \texttt{suransam@in.ibm.com}}
\author{Sameep Mehta}
\affil{IBM India Research Lab (IRL), New Delhi, India, \texttt{sameepmehta@in.ibm.com}}

\date{}

\maketitle

\begin{abstract}
Adversarial samples are strategically modified samples, which are crafted with the purpose of fooling a classifier at hand. An attacker introduces specially crafted adversarial samples to a deployed classifier, which are being mis-classified by the classifier. However, the samples are perceived to be drawn from entirely different classes and thus it becomes hard to detect the adversarial samples. Most of the prior works have been focused on synthesizing adversarial samples in the image domain. In this paper, we propose a new method of crafting adversarial text samples by modification of the original samples. Modifications of the original text samples are done by deleting or replacing the important or salient words in the text or by introducing new words in the text sample. Our algorithm works best for the datasets which have sub-categories within each of the classes of examples. While crafting adversarial samples, one of the key constraint is to generate meaningful sentences which can at pass off as legitimate from language (English) viewpoint. Experimental results on IMDB movie review dataset for sentiment analysis and Twitter dataset for gender detection show the efficiency of our proposed method.
\end{abstract}

\section{Introduction}
\label{sec:introduction}
Machine Learning (ML) algorithms can be deployed as a service (MLaaS) \cite{mlaas} where a trained model is available to the users for the task of classification/ regression of their query data. Applications like automatic chatbots, information retrieval (IR) systems uses ML algorithms of various complexities in the background. As the data is ever expanding, it is important for the deployer to update these ML models using the recently available data or the recent query result so that the overall performance of the ML model does not degrade on the newly available data. On the other hand, an adversary or an attacker aims to degrade the performance of the ML model. One way of doing that is to introduce some malicious queries to the model, which apparently seems to be a regular query sample to the deployer. The ML model gets confused by the malicious query sample and classifies it wrongly to other classes. As the classification performance degrades, the deployer tries to tune the parameters of the classifiers as per the feedback from the query samples (which may contain a large set of malicious adversarial samples) and eventually degrades the performance of the classifier even when tested against ordinary query samples. 

An adversarial sample \cite{kurakin16,adv_vae} can be defined as one which appears to be drawn from a particular class by humans (or advanced cognitive systems) but fall into a different class in the feature space. One way to prevent the attacks of the adversary is to train the classifier beforehand with the potential malicious samples \cite{goodfellow14}, we are assuming that there is only one type of adversarial samples used in the attack and limiting the problem setting to a single malicious data generation. We are considering Convolutional Neural Network (CNN) as our choice of classifier, which has been shown to produce good classification accuracy for the task of text classification.

With the advent of Deep Neural Network (DNN), classification of highly complex unstructured data has achieved good performance accuracy. Most of the time, DNNs are chosen as the most preferred classifier for image and text classification tasks. The highly complicated structure of the neural nets helps to model the intricacies of the complex features present in an complex data like image or a free flowing text document. There has been lots of studies to find out the vulnerabilities in a DNN \cite{huang}, and fool the classifier successfully.

Majority of the prior works in the field of synthesizing adversarial samples consider images data to work with \cite{goodfellow14,papernot16,gradient_masking,kurakin16}. Any value in the range of 0 to 255 can be interpreted as a valid pixel of an image. A minor change in the pixel values of an image, does generate a meaningful image and also the change is negligible to human eyes. These properties make the synthesize of a new image or modification of an image much easier, when compared to a more structured data like text. On the other hand, text samples are hard to modify or synthesize. Word2Vec \cite{word2vec} approach is a popular method used for data pre-processing and feature extraction for text samples. But here also, the discreet nature of the whole Word2Vec set makes it difficult to map any arbitrary vector to a valid word in the vocabulary. Another important aspect to generate adversarial text sample is to maintain the syntactic meeting of the sample along with the grammar, so that the changes remain difficult to detect for a human. All these challenges, makes the problem of adversarial text crafting very challenging and there has been little work \cite{adv_text} for adversarial text crafting in the text domain.

The proposed method of crafting adversarial samples in the text domain, is best suited for datasets having sub-categories. In this paper, we experiment with the IMDB movie review dataset \cite{imdb} extensively, for the task of sentiment analysis. The dataset can be sub-categorized using the genre of the movies considered for the reviews. Generally there is a set of high contributing words which increases the probability of belonging to a particular sentiment class of the reviews. For example, words like \textit{'good'}, \textit{'excellent'}, \textit{'like'} etc. indicates a positive review irrespective of the genre of the movie. However, there exists some words which are specific to the genre of the movie. For example: consider the sentence "\textit{The movie was hilarious}". This indicates a positive sentiment for a comedy movie. But the same sentence denotes a negative sentiment for a horror movie. Thus, the word \textit{'hilarious'} contributes to the sentiment of the review based on the genre of the movie. The classifier used for sentiment analysis neglect the genre information and determines the sentiment of the review text globally. However, using the sub-category level information, we can successfully create adversarial text samples. Another way to alter a review sample is to consider the synonyms or possible typos of high contributing words, which decreases the probability of belonging to the correct class. 

Often presence of adverbs can alter the sentiment of a review text. For example: consider the sentence: "The movie was fair". This sentiment may belong to either of the class. However, when we add the adverb 'extremely' and change the sentence to "The movie was extremely fair", it indicates that the movie was poor and should be treated as a negative sentiment. If, for the classifier, the word "extremely" do not contribute much to the probability of belongingness to a particular class, then the modified sample becomes an adversarial one. Similarly, removal of certain words may impact the class probability of the sample text in an opposite direction. 

The main contribution of this work is to craft adversarial samples in the domain of text data by preserving the semantic meaning of the sentences as much as possible. Also we try to minimize the alteration of the original input sample. To our best of knowledge, this is work aims to craft adversarial text samples with minimum possible modifications and preserves the semantic meaning of the sentence to the best possible extent. The rest of the paper is as follows: Section \ref{sec:related_work} describes the related work in the field of adversarial sample crafting, section \ref{sec:proposed_method} describes the proposed method and section \ref{sec:result} describes the experimental results. finally section \ref{sec:conclusion} concludes the paper.

\section{Related Work}
\label{sec:related_work}
Learning and deployment of machine learning models provides an opportunity for the hackers to exploit various sensitive informations about the features, nature of decision boundary etc. They may try to deteriorate the model by introducing false input or feedback to the system in this process. One way to confuse the classifier and deteriorate its performance is to introduce samples which apparently seems to be innocent by humans. These adversarial samples are hard to catch by the humans but causes the classifier to fail by a significant amount, due to which the classifiers may get falsely retrained. To resist the attack, the deployer can pre-assume some adversarial samples and generate the synthetic ones. Training the classifier with the synthetic adversarial samples with correct class-label makes it robust to the attack that can happen in future. We briefly talk about the techniques that are popular for creating adversarial samples.

Szegedy et al \cite{szegedyZSBEGF13} shows that the smoothness assumption of the kernels are not correct which makes the input-output mapping discontinuous in deep neural network. The adversarial samples are a results of this discontinuity, which lies in the pockets of the manifold of the DNNs. Thus the adversarial samples form the hard negative examples even though they lie in the distribution of the inputs provided to a DNN. A simple optimization problem that maximizes the network's prediction error is good enough to create adversarial samples for images.

\begin{figure}[htb]
\begin{center}
\includegraphics[width=8cm,height=4.5cm]{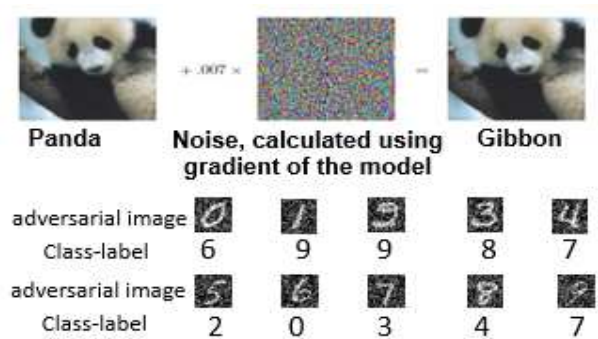}
\caption{Adversarial samples produced using Fast gradient Sign Method (FGSM) \cite{goodfellow14}, as predicted by trained DNN.}
\label{fig:fgsm}
\end{center}
\end{figure}

One of the most popular method of creating adversarial sample images from existing ones is by considering the gradient of the cost function of the classifier, which was introduced by Goodfellow et al \cite{goodfellow14}, and the resultant image becomes an adversarial sample. The noise is generated with the help of the gradient of the cost function of the classifier, with respect to the input image. This is a fast and effective way to create adversarial samples, which is popularly known as Fast Gradient Sign Method (FGSM). Figure \ref{fig:fgsm} shows how a panda image is modified to an adversarial sample which is now being classified as gibbon, as well as the adversarial samples along with their corresponding predicted class-labels from MNIST dataset \cite{lecun}, the classifier gets more generalized and it is less susceptible to adversarial attacks.

Papernot et al. \cite{gradient_masking}, an attacker can successfully generate adversarial samples. The attacker can train his own model using similar input data and create adversarial samples using FGSM method. These adversarial samples can confuse the deployed classifier also with good probability. Thus the attacker can be successful with almost no information of the deployed model. Papernot et al introduced the term 'gradient masking' and show its application on real world images like traffic signs, where the algorithm performs well.

Liang et al \cite{adv_text} proposed an algorithm to produce adversarial text samples using character level CNN \cite{character_cnn}. It shows the problem of directly using algorithms like FGSM for crafting adversarial text samples. The results were gibberish texts, which can be easily identified by humans as noisy samples. In their work, the important phrases are determined using back propagation of the gradient of the cost function. These phrases are replaced by pre-identified phrases from the training corpus, which are shown to change the class probability values. Three methods: insertion, deletion and modification, have been used for modifying the text to create an adversarial sample. However, detection of correct phrases and using them to create adversarial samples requires some heuristics, which are not very clearly mentioned in the paper. A codebase \cite{textfool} has been shared to craft adversarial text sample, along these ideas, which mostly modify existing words in the text sample by its synonyms and typos. Apart from the above mentioned work, there has been couple of notable works in the area of adversarial sample crafting in the text domain. Hossein Hossein et al \cite{google} shows that strategical insertion of punctuations with some selected words can fool a classifier and toxic comments can be bypassed when a model is used as a filer. However, the modified texts are easily detectable by humans and may not qualify as a good example of adversarial sample. Lastly, we can conclude that crafting of adversarial samples in text domain is a niche area when compared to its image counterpart.

\section{Proposed Method}
\label{sec:proposed_method}
We propose three different kinds of modifications to alter a regular input into an adversarial sample: (i) replacement, (ii) insertion and (iii) removal of words into the text. We aim to change the class-label of the sample by minimum number of alteration. The pseudo-code of the proposed method is given below in Algo. \ref{alg:algo1}.
\begin{algorithm}[htb]
\caption{Convert the input text $s$ into an adversarial sample}
\begin{algorithmic}[1]
\REQUIRE Sample text - $s$, Classifier trained for sentiment analysis $F$
\STATE Find class-label $y$ of the sample $s$: $y \leftarrow F(s)$
\STATE Find contribution $\mathcal{C}_F(w_i,y)$ of each word $w_i$ towards determining the class-label of the sample $s$ with respect to the classifier $F$.
\STATE Rank words according to $\mathcal{C}_F(\textbf{w},y)$: $\textbf{w} \rightarrow \{ w_1, w_2, \cdots, w_n\}$ where, $\mathcal{C}_F(w_1,y) >\mathcal{C}_F(w_2,y)> \cdots >\mathcal{C}_F(w_n,y)$
\STATE $i \leftarrow 1$
\WHILE{$y$ does not change}
\IF{$w_i$ is an Adverb and $\mathcal{C}_F(w_i,y)$ is considerably high}
\STATE Remove $w_i$ from $s$
\ELSE
\STATE Consider a candidate pool $\mathcal{P} = \{p_k\} \forall k$ for $w_i$
\STATE $j \leftarrow \underset{k}{\mathrm{argmin}} \mathcal{C}_F(p_k,y)$, $\forall p_k \in \mathcal{P}$
\IF{$w_i$ is Adjective and $p_j$ is Adverb}
\STATE Add $p_j$ before $w_i$ in $s$
\ELSE
\STATE Replace $w_i$ with $p_j$ in $s$
\ENDIF
\ENDIF
\STATE $y \leftarrow F(s)$
\STATE $i \leftarrow i+1$
\ENDWHILE
\end{algorithmic}
\label{alg:algo1}
\end{algorithm}
The steps of the algorithm are explained as below:

\subsection{Calculate contribution of each word towards determining class-label}
\label{subsec:word_contribution}
We target the words in the text in the decreasing order of their importance or contribution to the class belongingness probability. A word in the text is highly contributing if its removal from the text is going to change the class probability value to a large extent. Hence the contribution of a word $w_k$ can be measured as:
\begin{equation}
\mathcal{C}_F(w_k,y_i)= 
\begin{dcases}
    p_F(y_i|s) - p_F(y_i|s^{|w_k}), \text{if $F(s)=F(s^{|w_k})=y_i$} \\
    p_F(y_i|s) + p_F(y_j|s^{|w_k}),  \text{if $F(s) = y_i$ and $F(s^{|w_k})=y_j$}
\end{dcases}
\label{eq:word_contribution}
\end{equation}
where, $p_F(y_i|s)$ is the posterior probability of a sentence belonging to the class $y_i$ according to classifier $F$ and $s^{|w}$ denotes the sentence without the target word. However, for a large text sample, which consists multiple paragraphs, calculation of $\mathcal{C}_F(y_i|\textbf{w})$ is a time consuming process for each and every word. We use the concept of FGSM \cite{goodfellow14} to approximate the contribution of a word $w_k$, which can be calculated as:
\begin{equation}
\mathcal{C}_F(w_k,y) = -\nabla_{w_k} J(F,s,y_i)
\label{eq:fgsm}
\end{equation}
where, $y_i$ is the true class of the sentence $s$ containing the word $w_k$ and $J$ is the cost function of the classifier $F$ in hand: $F(s) = y_i$. Since, $-\nabla_sJ(F,s,y_i)$ denotes the adjustment to be made in the input $s$, to obtain the minimum cost function during training, it is a good way to determine the importance of each of the features for the particular classifier $F$. 

\subsection{Build candidate pool $\mathcal{P}$ for each word in sample text}
\label{subsec:candidate_pool}
We modify the sample text by considering each word at a time, in the order of the ranking based on the class-contribution factor. For this we consider synonyms and typos of each of the words as well as the genre or sub-category specific keywords which are briefly described as below.

\textbf{Synonyms and typos}: For each word, we build a candidate pool, which consists of the words with which the current word can be replaced with. For example: the word 'good' in the sentence 'The movie was good.' can be replaced with 'nice', 'decent' etc. Here we are considering the synonyms of the word 'good'. Another possible way to extend the candidate pool is to consider the possible typos that may happen while typing the review. However, if we consider too many typos the input text $s$ can turn into a meaningless piece of text. Also the typos attract human attention quiet a lot and it can be easily detectable by any spell checker. Hence, we consider typos which are valid words. For example, for the word 'good', we will consider the typos 'god', 'goods' etc, which are valid English words. 

\textbf{Genre specific keywords:} Apart from using synonyms and typos, we also consider some sub-category or genre specific keywords. This is based on the fact that certain words may contribute to positive sentiment for a particular genre but can emphasis negative sentiment for other kind of genre of movies. These keywords capture the distinctive properties of the classes by considering the term frequencies (tf) in the corpus. If the tf of a word is high for a particular class of sample texts and is low for texts belonging to a different class, then we can safely say that the work is distinctive for the first class. Let, $\delta_i$ denote the set of distinctive keywords for the $i^{th}$ class. Since we are dealing with two class problem, let the set of distinctive words for class 1 and 2 be denoted as $\delta_1$ and $\delta_2$ respectively. Now, to consider the sub-category or the genre information, we are considering the set of distinctive words for the two classes from the texts belonging to each of the genres separately. Let these sets be denoted as $\delta_{1,k}$ and $\delta_{2,k}$ for the two classes of samples for the particular genre $k$. Now, using these sets of keywords, we add terms in the candidate pool as:
\begin{equation}
\mathcal{P} = \mathcal{P} \cup \{ \delta_j \cap \delta_{i,k} \}
\label{eq:genre_keywords}
\end{equation}
where, $i,j \in \{ 1,2\}$, $F(s) = y_i$ and $i \neq j$ and $k$ is the genre class of the word for which we are building the candidate pool $\mathcal{P}$ of words. Here, we are considering a subset of candidate words which are  distinctive to the opposite class when the sub-category information is being avoided. So each of these words will contribute negatively towards the contribution of the text sample for the $i^{th}$ class and are ideal for crafting an adversarial sample. 

\subsection{Crafting the adversarial sample}
\label{subsec:adv_crafting}
We consider three different approaches to change the given text sample $s$ at each iteration, so that the modified sample $s'$ flips its class label. Let at any iteration, the word chosen for modification is denoted as $w_i$. Then the heuristics for modifying the text $s$ are explained as below:

\textbf{Removal of words:} At first we check if the word under consideration i.e. $w_i$ is an adverb or not. If $w_i$ is an adverb and its contribution score $\mathcal{C}_F(w_i,y)$ is considerably high, then we remove the word $w_i$ from $s$ to get the modified sample $s'$. The motivation behind this heuristics is that adverbs put an emphasis on the meaning of the sentences, and often do not alter the grammar of the sentence by its introduction or removal from a sentence.

\textbf{Addition of words:} If the first step of modification is not satisfied, we select a word from $\mathcal{P}$, the candidate pool of $w_i$, using the concept of FGSM. If the selected from from $\mathcal{P}$ is $p_j$, then the following condition is being satisfied:
\begin{equation}
j = \underset{k}{\mathrm{argmin}} \mathcal{C}_F(p_k,y) \,\,\, \forall k
\end{equation}
Now, if $w_i$ is an adjective and $p_j$ is an adverb, we modify $s$ by inserting $p_j$ just before $w_i$ to get $s'$.

\textbf{Replacement of word}: In case the conditions of the first two steps are not satisfied, we replace $w_i$ with $p_j$ in $s$ to obtain $s'$. Now, if $p_j$ is obtained from the genre specific keywords, then we consider it for replacement only if the parts of speech of both $w_i$ and $p_j$ are the same. Otherwise, we select the next best word from $\mathcal{P}$ and do the replacement. The matching of the parts of speech is necessary to avoid detection of the modified sample $s'$ by the humans as an adversarial sample. Also, this condition ensures that the grammar of the sentence does not gets corrupted too much.

We keep on changing each word at a time in $s$ to obtain $s'$, unless the class-label of $s$ and $s'$ becomes different. Since we consider words in the order of their contribution score $\mathcal{C}_F(w_i,y)$, we are crafting adversarial samples in the least possible changes using the idea of greedy method. The minimum number of changes in the sample text $s$, ensures that the semantics and the grammar of the adversarial sample $s'$ remains similar to that of$s$. Also, the conditions that we enforced with the parts of speech of the words, helps us to maintain the structure of the sentence to a large extent.

Once we obtain the adversarial samples, we re-train our existing text classifier to make it more robust to adversarial attacks, as done in \cite{goodfellow14}.

\section{Experimental results}
\label{sec:result}
We show our experimental results on two datasets (i) IMDB movie review dataset \cite{imdb} for sentiment analysis and (ii) twitter dataset for gender classification \cite{twitter}. We compare our method with the existing method, TextFool, as described in \cite{adv_text} using Twitter dataset. We implemented the method described in \cite{adv_text} to the best of our abilities, but it may not be an exact implementation due to lack of proper explanation of the heuristics mentioned in \cite{adv_text}. We evaluate the algorithms for adversarial sample crafting as done in \cite{goodfellow14}. The classification accuracies of the original and the tainted test data is shown on two types of model. The first one is trained on the original training set as obtained from the respective datasets. The second model is obtained after re-training the first one using adversarial samples obtained by modifying the samples from the original training set. Classification accuracy for a dataset can be defined as the percentage of the samples correctly labeled by the classifier in hand, where the correct label is a pat of the database. Detailed study of the experiments on these two datasets are given below.

\subsection{IMDB movie review sentiment analysis}
\label{subsec:imdb}
The IMDB movie review dataset consists of the reviews for different movies along with the class-label (positive or negative sentiment) and the url from which the review has been taken. The dataset is divided into training and testing sets, each of with contains 22500 positive samples and 22500 negative samples.

\textbf{Data preprocessing and feature extraction:} We manually found the genre of the movie by using the IMDB api called IMDBpy using the url of the given reviews. Next, we identified the genre of movies having considerable amount of reviews in the dataset. After filtering, we are considering the reviews from movies having genres 'Action', 'Comedy' or 'Drama'. These genres are considered as sub-categories which are used for selecting genre specific distinctive keywords. Finally, we extract the features using the toolbox spacy \footnote{https://spacy.io/}, where features denote the number of times the word is occurring in the text sample.

\begin{figure}[htb]
\begin{center}
\includegraphics[width=9.9cm,height=3.3cm]{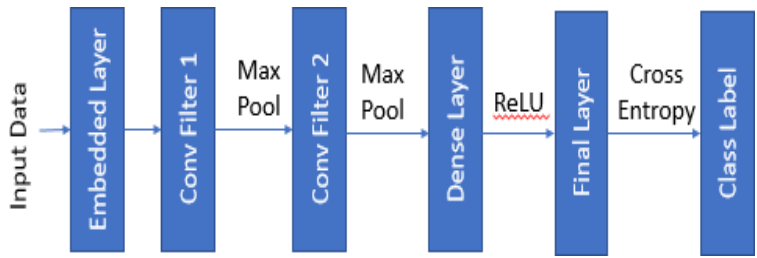}
\caption{Architecture of the CNN used for sentiment classification of IMDB movie review dataset.}
\label{fig:imdb_cnn}
\end{center}
\end{figure}

\textbf{Classifier Used:} We have used Convolutional Neural Network (CNN) for classification purpose. The block diagram of the CNN architecture is shown in Fig. \ref{fig:imdb_cnn}. 

\textbf{Results:} We show the efficiency of our proposed method by comparing the accuracy of the model obtained at different configurations. Another important factor for evaluating the effectiveness of adversarial samples is to measure the semantic similarity between the original samples and their corresponding tainted counterparts. A less similarity score denotes that the semantic meaning of the original and the modified samples are quite different, which is not desired. The number of changes incurred to obtain the adversarial sample is also a good measure for evaluating the algorithm. The number of changes made to craft a successful adversarial sample should be ideally low. It is also a measure which indicates that the words contributing highly towards the determination of the class-label (words with high value of $\mathcal{C}_F(w,y)$, see Eqn. \ref{eq:fgsm}) are detected and replaced earlier in the process of crafting. 

Unlike images, a text sample may not be always convertible to its adversarial counterpart. This is due to be presence of the constraints used in the crafting process, which are necessary for building a meaningful text sample. This is also observed while evaluating the method Textfool \cite{adv_text}, for comparative study. Thus, the semantic similarity between a text sample and its adversarial counterpart is only valid when the adversarial sample has been crafted successfully.

In our experimentation, we use the training and the test set as given in the IMDB dataset explicitly. This accuracy is the baseline (third row in Table \ref{tab:imdb}), with which we compare other variations considered in the experimentation. We find the adversarial counterpart of the samples in the test set and compare it with the baseline accuracy (forth row in Table \ref{tab:imdb}). The above mentioned measures are given in Table \ref{tab:imdb}. In order to show the effectiveness of using genre specific keywords in the candidate pool during the process of crafting, we show all the above mentioned values in Table \ref{tab:imdb} when the genre specific keywords were not considered in the candidate pool (forth column in Table \ref{tab:imdb}). Apart from the model performance, we also observe the number of test samples converted to their corresponding adversarial samples successfully (fifth row in Table \ref{tab:imdb}), and the semantic similarity between these pairs. The semantic similarity is measured using Spacy toolbox. The average semantic similarity between the original text sample and their adversarial counterparts (for test set only) are 0.9164 and 0.9732 with and without using the genre specific keywords respectively. The semantic similarity between the original and their corresponding perturbed samples does decrease a bit while the genre specific keywords in the candidate pool, but the number of valid adversarial samples generated also increases in this case (see $5^{th}$ row of Table \ref{tab:imdb}). The seventh and the eighth row in Table \ref{tab:imdb} show the classification accuracy of the original test set and the perturbed test set after re-training the CNN with adversarial samples. We can see that the difference in the accuracies in these two rows are very less, which shows that the model has generalized well after retraining. 

\begin{table}[!h]
\centering
\begin{tabular}{|p{3cm}|p{3cm}|p{3cm}|p{3cm}|}
\hline
& TextFool \cite{adv_text} &{Proposed method using genre specific keywords} & {Proposed method w/o using genre specific keywords}\\\hline \hline
\multicolumn{4}{|c|}{CNN trained with original training set} \\ \hline
Accuracy using original test set & 74.53 & 74.53 & 74.53 \\ \hline
Accuracy using adversarial test set & 74.13 & 32.55 & 57.31 \\ \hline
Percentage of perturbed samples & 0.64 & 90.64 & 42.76 \\ \hline
\multicolumn{4}{|c|}{CNN re-trained with perturbed training set} \\ \hline
Accuracy using original test set & 68.14 & 78.00 & 78.81 \\ \hline
Accuracy using adversarial test set & 68.08 & 78.46 & 78.21 \\ \hline
\end{tabular}
\caption{Performance results on IMDB movie review dataset.}
\label{tab:imdb}
\end{table}

\begin{figure}[!b]
\begin{center}
\includegraphics[width=8cm,height=6cm]{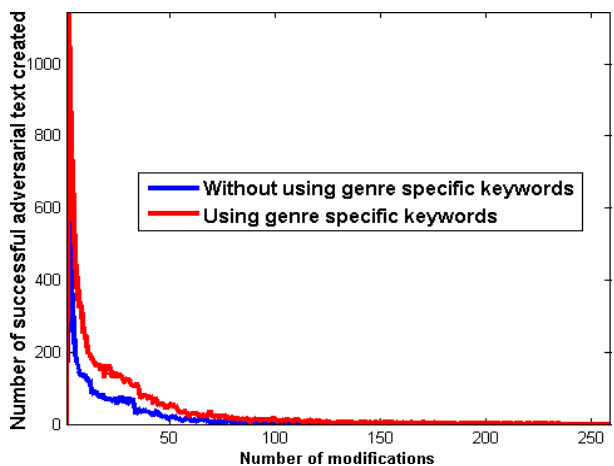}
\caption{Plot showing the number of adversarial samples produced against the number of changes incurred in the text samples (i) with (red curve) and (ii) without (blue curve) using genre specific keywords in the IMDB movie review dataset.}
\label{fig:imdb_chart}
\end{center}
\end{figure}

From Table \ref{tab:imdb} it is evident, that the proposed method of adversarial sample crafting for text is capable of synthesizing semantically correct adversarial text samples from the original text sample. The table also shows that the inclusion of genre specific keywords definitely boost up the quality of sample crafting. This is evident from the fact the drop in accuracy of the classifier before re-training for original text sample and the adversarialy crafted text sample is more when genre specific keywords are being used. Figure \ref{fig:imdb_chart} shows a graph that indicates the number of changes required to create successful adversarial samples with and without using the genre-specific keywords. We can observe that when we use genre specific keywords for creating adversarial samples, the number of tainted sample produced is more than that when genre specific keywords are used, for same number of changes done in two cases (red curve always above blue curve).

\subsection{Twitter data for gender prediction}
\label{subsec:twitter}
This dataset \cite{twitter} contains twitter comments on various topics from male and female users, and the task is to predict the gender by looking at the tweet data. The dataset contains 20050 instances and 26 features. Though the twitter dataset is much smaller in size and the tweets are also shorter in length than the reviews in the IMDB dataset, but it is very challenging as the tweets contain lots of urls and hash tag descriptions.

\textbf{Data preprocessing and feature extraction:} We consider two features of the dataset, 'texts' and 'description', for determining the gender. Out of 20050 instances, we only consider the instances where the feature 'gender confidence' score is 1. This leaves us with 10020 samples, out of which 8016 samples are used for training and 2004 number of samples are used for testing purpose. We concatenate the strings from the features 'text' and 'description' and do the feature extraction for the classification task. The number of occurrence of each of the words in each of the tweets (text and description) have been considered as the features in this experimentation. The average semantic similarity between the text samples and their corresponding adversarial counterpart in the test set are 0.972 and 0.99 for our proposed method and Textfool respectively. Hence we can say that our proposed method has been able to create large number of semantically correct adversarial samples for Twitter dataset.

\begin{figure}[htb]
\begin{center}
\includegraphics[width=9.9cm,height=3.3cm]{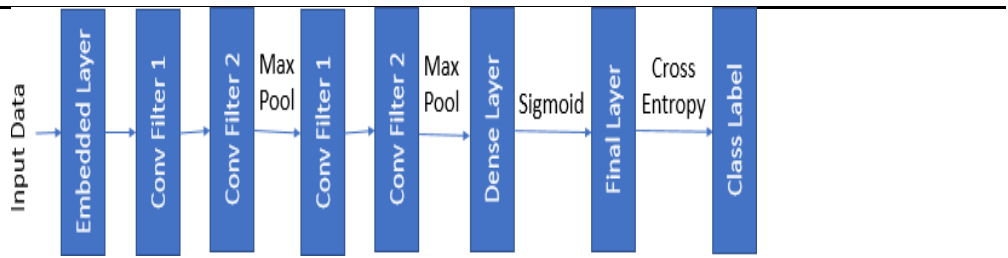}
\caption{Architecture of the CNN used for Twitter gender classification dataset.}
\label{fig:twitter_cnn}
\end{center}
\end{figure}

\textbf{Classifier Used:} We have used CNN for gender prediction task for twitter dataset. The block diagram of the CNN architecture is shown in Fig. \ref{fig:twitter_cnn}. 

\textbf{Results:} We show the efficiency of our proposed method as done for the IMDB movie review dataset. For this dataset, we compare our work with that of the performance of TextFool \cite{adv_text}. Table \ref{tab:twitter} shows the classification accuracy when the classifier is trained with two different training dataset. In the first case, the classifier is trained with original train samples obtained by spiting the dataset, as mentioned above (third, forth and fifth row in Table \ref{tab:twitter}). Next, we synthesize adversarial samples from the training set and use those to retrain the classifier and observe classification accuracies on clean and tainted text set respectively (seventh and eighth row in Table \ref{tab:twitter}). Table \ref{tab:twitter} shows the classification accuracy of the original and tainted text samples before and after re-training the model. It also shows the number of successful adversarial samples created from the test set (fifth row). It is evident that our method not only has been able to produce a larger number of adversarial samples, but they are also close in terms of semantic meaning of the text from their original form. 
 
\begin{table}[htb]
\centering
\begin{tabular}{|p{3cm}|c|c|}
\hline
&{\bf TextFool} & {\bf Proposed method}\\\hline \hline
\multicolumn{3}{|c|}{CNN trained with original training set} \\ \hline
Accuracy using original test set & 63.43 & 63.43 \\ \hline
Accuracy using adversarial test set & 63.23 & 50.10 \\ \hline
Ratio of perturbed samples & 0.03 & 58.53 \\ \hline
\multicolumn{3}{|c|}{CNN re-trained with perturbed training set} \\ \hline
Accuracy using original test set & 61.82 & 61.68 \\ \hline
Accuracy using adversarial test set & 61.34
& 60.07 \\ \hline
\end{tabular}
\caption{Performance results on Twitter gender classification dataset.}
\label{tab:twitter}
\end{table}

\begin{figure}[htb]
\begin{center}
\includegraphics[width=14.1cm,height=10.2cm]{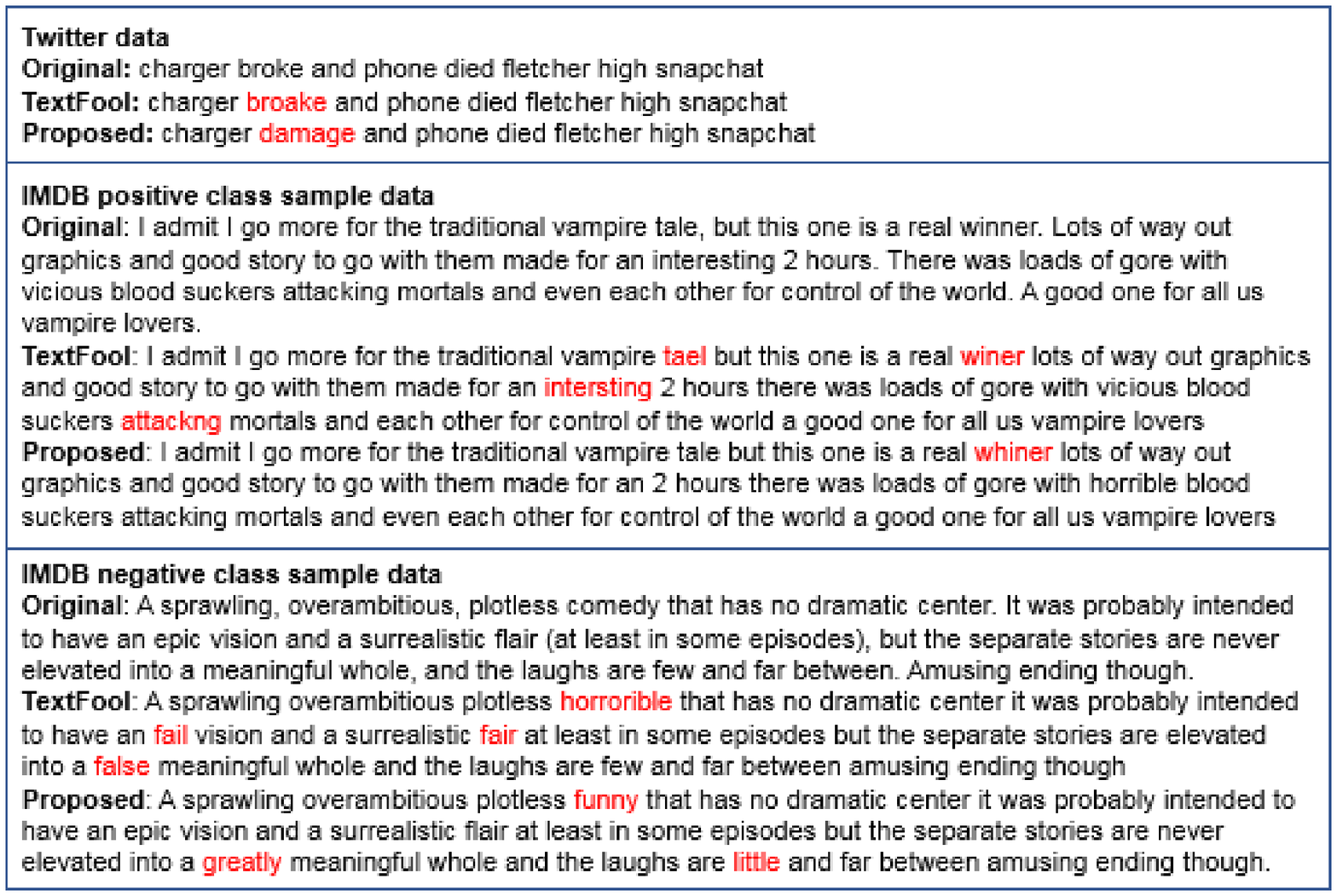}
\caption{Examples of adversarial samples crafted from Twitter and IMDB dataset using (i)TextFool and (ii) proposed method.}
\label{fig:example}
\end{center}
\end{figure}

Finally we show some of the examples of synthetic adversarial samples produced using TextFool and our proposed method from both the datasets in Fig. \ref{fig:example}.The examples show examples of different rules (insertion, replacement and deletion of words) incorporated in our proposed method

\section{Conclusion}
\label{sec:conclusion}
In this paper we describe a method to modify an input text to create an adversarial sample. To our best of knowledge, this is one of the first papers to synthesize adversarial samples from complicated text samples. Unlike images, in texts, a number of conditions must be satisfied while doing the modification steps to ensure the preservation of semantic meaning and the grammar of the text sample. In this paper, we are considering each word at a time and selecting the most appropriate modification in a greedy way. However, a much better approach is to consider each of the sentences in the text sample and modifying it which eventually confuses the classifier. In this work, the steps adopted for modifications are heuristic in nature, which can be improved and automated further to obtain better results.

\bibliographystyle{abbrvnat}
\bibliography{list_of_papers}

\end{document}